\documentclass[letterpaper, 10 pt, conference]{ieeeconf}  

\IEEEoverridecommandlockouts                              

\makeatletter
\let\NAT@parse\undefined
\makeatother
\usepackage[numbers,sort&compress]{natbib}

\usepackage[pdftex]{graphicx}
\usepackage{amsmath}
\usepackage{amssymb}
\usepackage{subcaption}
\usepackage{multirow}
\usepackage{array,booktabs}
\usepackage{diagbox}
\usepackage{balance}
\usepackage{xspace}
\usepackage{algorithm}
\usepackage{algpseudocode}
\usepackage{adjustbox}
\usepackage{romannum}
\usepackage{mathtools}
\usepackage{array}

\usepackage[final]{hyperref}
\hypersetup{
    colorlinks=true,
    linkcolor=magenta,
    filecolor=magenta,
    urlcolor=magenta,
    citecolor=magenta
}

\usepackage{cleveref}
\crefname{figure}{Fig.}{Figs.}
\Crefname{figure}{Fig.}{Figs.}

\usepackage[font=small]{caption}

\usepackage{rpm_acronyms}
\usepackage{rpm_misc}

\usepackage{soul,color}
\usepackage{lipsum}

\usepackage{xcolor}

\usepackage{tikz} 
\usepackage{todonotes}

\title{\LARGE \bf

SG-ODOM: Event-Driven Proprioceptive State Estimation with Stability-Gated Contact Updates for Legged Robots

}

\author{Francesco Iotti$^{1}$$^{*}$, Chiyun Noh$^{3}$$^{*}$, Franco Angelini$^{1}$, Manolo Garabini$^{1}$,\\
Ayoung Kim$^{3}$, Maurice Fallon$^{2}$, and Frank Dellaert$^{4}$%
\thanks{
$^{*}$These authors contributed equally to this work. \newline
This document is the result of the research project funded by the Italian Ministry of University and Research (MUR) - Fondo Italiano per la Scienza Applicata (FISA), through the OCCAM project under Grant FISA 2023-00324 CUP I53C25000700001 and the ``FoReLab'' (Future-oriented Research Lab) Project (Departments of Excellence), and partially by the European Union HORIZON-MSCA-2023-SE-01-01 - MSCA Staff Exchanges 2023 Program under the Grant Agreement No. 101182891 (NEUTRAWEED) and the UK EPSRC project Mobile Robot Inspector (UKRI9508). For the purpose of open access, the authors have applied a Creative Commons Attribution (CC BY) license to any Accepted Manuscript version arising. F. Iotti is within the Italian Doctorate of Robotics and Intelligent Machines.\newline
Corresponding author: Francesco Iotti,\newline
{\tt\small (francesco.iotti@phd.unipi.it)}.}
\thanks{$^{1}$Centro di Ricerca ``Enrico Piaggio'' and Dipartimento di Ingegneria dell'Informazione, Universit\`a di Pisa, Pisa, Italy.}%
\thanks{$^{2}$Dept. of Engineering Science, University of Oxford, Oxford, UK.}%
\thanks{$^{3}$Department of Mechanical Engineering, Seoul National University, Seoul, Republic of Korea.}%
\thanks{$^{4}$College of Computing, Georgia Inst. of Technology, Atlanta, GA, USA.}%
}

\begin{document}
\bstctlcite{IEEEexample:BSTcontrol}
\maketitle
\thispagestyle{empty}
\pagestyle{empty}

\begin{abstract}

Reliable proprioceptive state estimation for legged robots fuses inertial sensing with kinematic sensing from leg encoders. The latter assumes a foot is stationary during the gate; however, physical foot touchdown does not necessarily provide a stationary reference for odometry, particularly during impact and load transfer. Repeatedly applying kinematic corrections throughout the same contact phase can introduce redundant or unreliable information. We present SG-ODOM, a proprioceptive odometry framework that uses stability-gated contact selection to schedule kinematic updates. Following physical touchdown, a stability gate accepts the contact for state estimation only after the foot reaches a sufficiently stable support condition. The resulting validated touchdown event triggers an estimator correction, while additional timeout updates bound the maximum duration of uninterrupted IMU propagation. 
The framework is evaluated on multiple legged platforms, both quadrupedal and humanoid robots, against representative state-of-the-art proprioceptive estimators. Across the evaluated sequences, SG-ODOM demonstrates competitive or superior estimation accuracy, reducing \ac{RPE} by up to $25\%$ compared to the baseline estimator on outdoor sequences from the GrandTour dataset. We make our source code publicly available.
\footnote{Github Repo: \url{https://github.com/ori-drs/SG-ODOM}}
A supplementary video is also provided.\footnote{Video: \url{https://youtu.be/-583_RyuhqQ}}


\end{abstract}

\section{Introduction}
\label{sec:intro}

Reliable legged locomotion requires stable, low drift, low latency and high-frequency estimates of the robot's base motion. Proprioceptive odometry commonly uses the \ac{IMU} for high-rate state propagation and leg kinematics for state updates~\cite{bloesch2013state,hartley2019contact}. Leg kinematics help correct accumulated inertial drift by relating the robot base to feet in contact with the environment. Such indirect state measurements assume that the selected contact frame provides a stationary reference.

However, physical contact does not in itself ensure a stationary reference, particularly during touchdown and load transfer. A foot may sustain a substantial \ac{GRF} while still bouncing or settling on the ground~\cite{lin2022learnedcontact}. Contact information should therefore indicate whether the foot provides a stable reference for estimation.

Even when a contact provides a stable reference, it remains unclear whether kinematic corrections need to be applied at the full joint-sensing rate. Consecutive measurements within the same contact phase are strongly correlated and may therefore provide limited additional information. Moreover, frequent kinematic corrections can repeatedly introduce errors caused by slip or residual foot motion. These errors can distort the estimated base motion and introduce spurious changes in the state estimate. Conversely, overly sparse corrections increase the duration of IMU-only propagation and allow inertial drift to accumulate. This raises the question of how frequently kinematic corrections should be applied relative to inertial propagation.

\begin{figure}
    \centering
    \includegraphics[width=\linewidth]{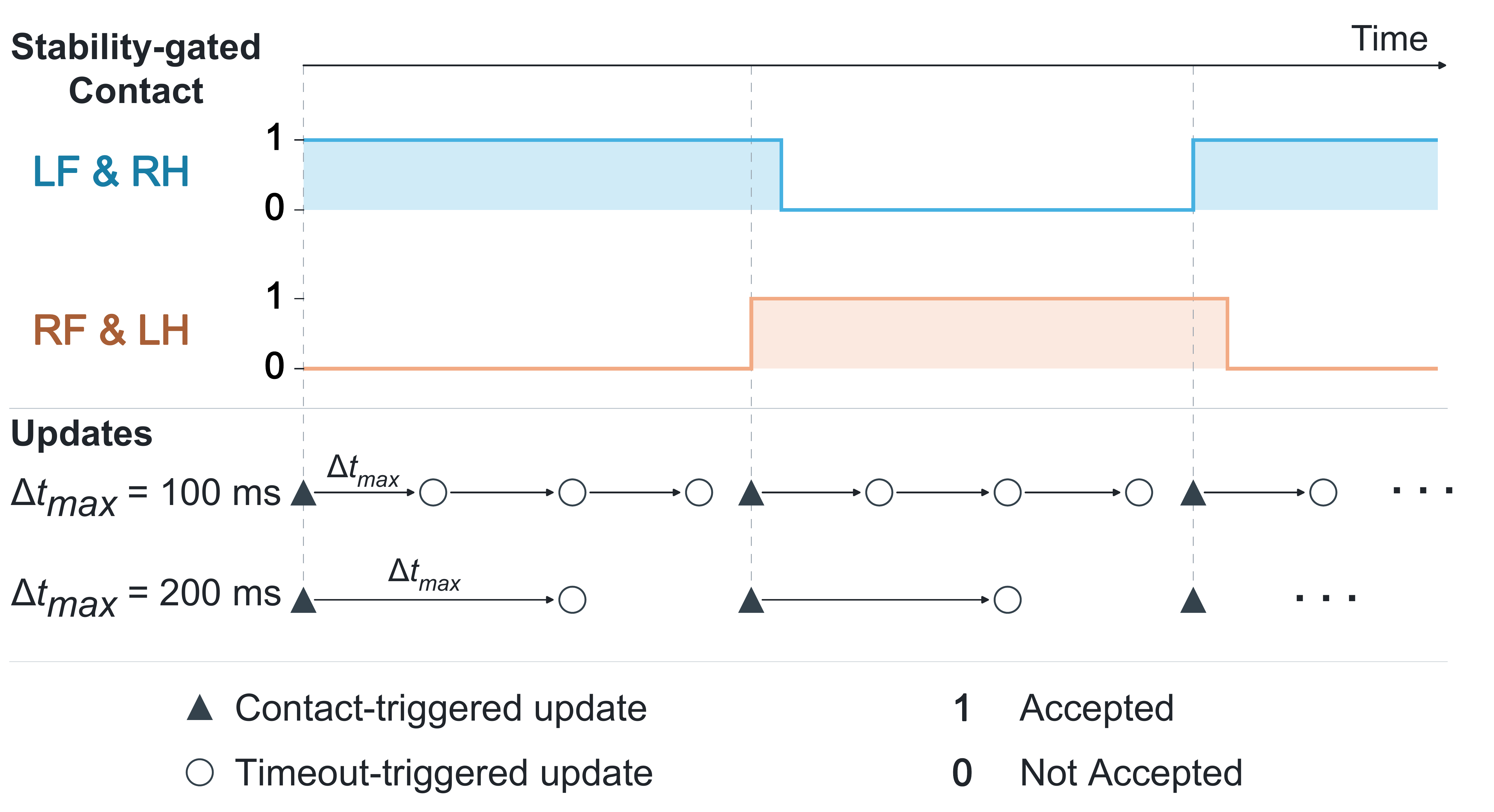}
    \caption{Illustration of the event-based contact update for a  quadruped trot gait. Stable touchdown events initiate immediate contact-triggered updates. Later timeout-triggered updates bound IMU-only propagation (below $\Delta t_{max}$). Each update resets the dead-reckoning interval, resulting in event-driven rather than fixed-rate updates. This is illustrated here for two different thresholds of 100~ms and 200~ms.}
    \label{fig:event-based-contact}
    \vspace{-7mm}
\end{figure}

In this work, we propose SG-ODOM, a stability-gated, event-driven proprioceptive odometry strategy that preserves high-rate IMU propagation and contact processing while selectively applying kinematic corrections. A physical touchdown is accepted for estimation only after the foot reaches a sufficiently stable contact condition, avoiding corrections during the transient phase following impact. These validated touchdowns initiate immediate contact-triggered updates, while, during unchanged contact conditions, timeout-triggered updates are applied at a lower rate than the joint-encoder measurements, but frequent enough to avoid prolonged IMU-only propagation, as illustrated in Fig.~\ref{fig:event-based-contact}.

We implement this proposed strategy using two estimator configurations: an \ac{InEKF}, and a fixed-lag smoother. Both configurations share the same sensing, contact, and update interfaces, allowing the effect of estimator formulation to be examined under consistent conditions. Robot-specific sensing and kinematic processing are handled in a separate front end, enabling the same estimation back end to be used across quadrupedal and humanoid platforms.

Our contributions are as follows:




\begin{itemize}
    \item An event-driven proprioceptive odometry strategy that
    separates high-rate IMU propagation and contact evaluation
    from kinematic correction timing, using validated touchdowns
    for immediate contact-triggered updates and timeout-triggered updates to
    limit IMU-only propagation.

    \item A stability-gated contact classifier that distinguishes
    physical touchdown from estimation-valid support, accepting
    contacts only after loading and force-stability conditions
    persist, past the impact transient.

    \item An evaluation on three legged platforms with ablations
    showing that stability gating improves accuracy for both
    estimators and that lower correction frequencies can preserve
    or improve accuracy. The fixed-lag configuration achieves
    approximately 25\% lower average translational RPE than the
    onboard ANYbotics estimator across seven GrandTour missions.
    
    \item An open-source GTSAM-based implementation with shared sensing,
    contact, and update interfaces supporting both an InEKF and
    a fixed-lag smoother, with a robot-specific front end.

\end{itemize}

The framework is evaluated on sequences from the GaRLILEO dataset \cite{noh2026garlileo} (Boston Dynamics Spot) and the GrandTour dataset \cite{frey_tuna2026grandtour} (ANYbotics ANYmal D), and on a self-collected dataset with the Unitree G1, as illustrated in Fig.~\ref{fig:cover_pic}. The evaluation examines the effect of contact classification and update frequency, and compares the two SG-ODOM configurations with existing proprioceptive and manufacturer-provided estimators.

\section{Related Works}
\label{sec:related:works}

\subsection{Proprioceptive State Estimation}

Existing approaches to proprioceptive state estimation for legged robots can be broadly categorized into filtering and optimization-based methods. Filtering-based estimators recursively propagate and correct the current state, whereas optimization-based approaches jointly estimate multiple states over a temporal window. Bloesch et al.~\cite{bloesch2013state} fused IMU measurements with leg kinematics in an \ac{EKF}-based contact-aided estimator. Hartley et al.~\cite{hartley2019contact} introduced the Contact-Aided \ac{InEKF}, exploiting Lie-group structure while explicitly handling contact addition and removal, while DRIFT~\cite{drift} extended invariant proprioceptive estimation across multiple robot platforms.

Among optimization-based approaches, factor-graph smoothers~\cite{Dellaert17fnt_factorGraphs, Dellaert21ar_factorGraphs} jointly incorporate inertial, kinematic, and contact-related measurements over a temporal window. Wisth et al.~\cite{wisth2019robust} combined inertial, leg, and visual odometry within a joint optimization framework. Fourmy et al.~\cite{fourmy2021contact} introduced contact-force preintegration, while Agrawal et al.~\cite{agrawal2023proprioceptive} modeled the complete kinematic chain together with IMU preintegration and contact information. Yoon et al.~\cite{yoon2024invariant} further proposed an invariant smoother that incorporates dynamic contact-event information and jointly optimizes inertial and contact measurements. Holistic Fusion~\cite{nubert2024holistic} addresses estimator portability through a task- and setup-agnostic factor-graph formulation.

While prior work has extensively studied estimator formulation and the modeling and treatment of kinematic and contact measurements, the timing of kinematic corrections within a persistent contact phase has received limited attention. Frequent updates may repeatedly incorporate correlated measurements, while sparse updates increase the duration of IMU propagation. The present work examines when kinematic corrections should be applied. Rather than updating at the leg-encoder sampling rate, we trigger corrections at validated contact events, while a maximum update interval bounds the duration of IMU-only propagation when no new contact event occurs.

\subsection{Contact Estimation and Handling}

Reliable contact-aided state estimation depends on accurately determining the contact state from proprioceptive measurements. Camurri et al.~\cite{camurri2017probabilistic} inferred contact from force measurements and robot kinematics through a probabilistic framework that accounted for impact detection. Maravgakis et al.~\cite{maravgakis2023probabilistic} estimated the probability of stable contact from inertial information to distinguish from physical contact, while Sun et al.~\cite{sun2025legstate} used a probabilistic model with proprioceptive feedback to estimate leg phases. Han et al.~\cite{han2023contact} combined multiple proprioceptive signals for contact detection over unstructured terrain.

 More recent approaches use learned models to infer contact state or contact probability from proprioceptive measurements. Gökbakan et al.~\cite{gokbakan2025contact} proposed a data-driven contact estimator designed to reduce dependence on dedicated contact sensing and gait-specific assumptions, while Youm et al.~\cite{youm2025legged} estimated contact probability with a neural measurement model and incorporated it into an InEKF-based state estimator. Xie et al.~\cite{xie2026lowdrift} combined a physics-informed contact estimation network with full joint-state information to reduce drift in proprioceptive state estimation.

Rather than estimating contact state itself, other works focus on how uncertain or imperfect contact information should influence the estimator. STEP~\cite{kim2022step} introduces a preintegrated foot-velocity factor rather than relying exclusively on a perfectly stationary contact point. Gu et al.~\cite{gu2024contact} proposed a learned contact bias to improve leg odometry, while Lee and Kim~\cite{lee2026attention} introduced a neural augmentation that compensates for slip-induced errors in the kinematic update. Adaptive invariant filtering approaches~\cite{kim2025adaptive} instead modify the uncertainty assigned to kinematic measurements according to the estimated contact condition, reducing the influence of unreliable updates.

Prior work has improved contact detection, estimated contact probability, and adapted the influence of uncertain kinematic measurements. However, physical contact does not necessarily imply that the foot has reached a sufficiently stable support condition for a kinematic correction, particularly during touchdown and load transfer. We address this distinction by accepting contact only after a stability gate is satisfied and using the resulting validated touchdown as an event for kinematic correction.

\section{Estimator Framework and Configurations}
\label{sec:implementation}


To support consistent evaluation across different robot platforms and estimator formulations, we adopt the front end/back end architecture used by other proprioceptive legged state estimators~\cite{camurri2020pronto}. A \emph{robot-specific front end} processes platform-dependent IMU and joint measurements, computes forward kinematics, and converts them into common inputs consisting of IMU measurements, foot positions, and contact information. A shared \emph{filtering \& smoothing back end} consumes only these inputs, independent of the robot platform or contact classifier that produced them (Fig.~\ref{fig:gtsam_configuration}).

This section describes this common estimation architecture and the two estimator configurations used to implement SG-ODOM: an \ac{InEKF} and a fixed-lag smoother. Both configurations share the same sensing inputs, contact model, and stability-gated event-driven update schedule, while differing in their estimation formulation and correction mechanism.



\subsection{Common State and Contact Model}
\label{sec:common-model}

We denote the robot body and navigation frames by $\mathcal{F}^{B}$ and $\mathcal{F}^{N}$, respectively. The floating-base quantities estimated by all configurations are collected as
\begin{equation}
    X_k =
    \left(
        R^N_{B,k},\,
        p^N_{B,k},\,
        v^N_{B,k}
    \right)
    \in \mathrm{SE}_{2}(3),
    \label{eq:event-state}
\end{equation}
where $R^N_{B,k}\in\mathrm{SO}(3)$ maps body to navigation frame, and $p^N_{B,k},v^N_{B,k}\in\mathbb{R}^{3}$ denote the base position and velocity in $\mathcal{F}^{N}$. Footholds and IMU biases are incorporated differently by the filtering and smoothing configurations.

For an estimator-valid contact of foot $i$, the corresponding foothold $f_i^N$ is modeled as stationary in $\mathcal{F}^{N}$. The foot position obtained from forward kinematics provides the common contact measurement, with time indices omitted for clarity,
\begin{equation}
    z_i^B
    =
    (R_B^N)^\top
    \left(f_i^N-p_B^N\right)
    + n_i,
    \label{eq:contact-model}
\end{equation}
where $z_i^B$ denotes the measured foot position expressed in $\mathcal{F}^{B}$, and $n_i$ denotes the measurement noise. The contact classifier determines which contacts are eligible for kinematic updates, without changing the underlying measurement model. The same physical contact model is used by all estimator configurations. The InEKF maintains the currently tracked foothold $f_i^N$ within the filter state, whereas the fixed-lag smoother represents each accepted contact episode by a distinct navigation-frame landmark $f_{i,e}^N$, where $e$ denotes the contact-episode index.

A new contact episode starts when a contact event is validated. At its first update, the corresponding foothold is initialized as
\begin{equation}
    \hat{f}^N_i
    =
    \hat{p}^N_B
    +
    \hat{R}^N_B z^B_i,
    \label{eq:foothold-init}
\end{equation}
using the current base estimate. Subsequent contact measurements constrain the same foothold  without reinitialization.

\subsection{Estimator Configurations}
\label{sec:estimator-configurations}

\begin{figure}
    \centering
    \includegraphics[width=\linewidth]{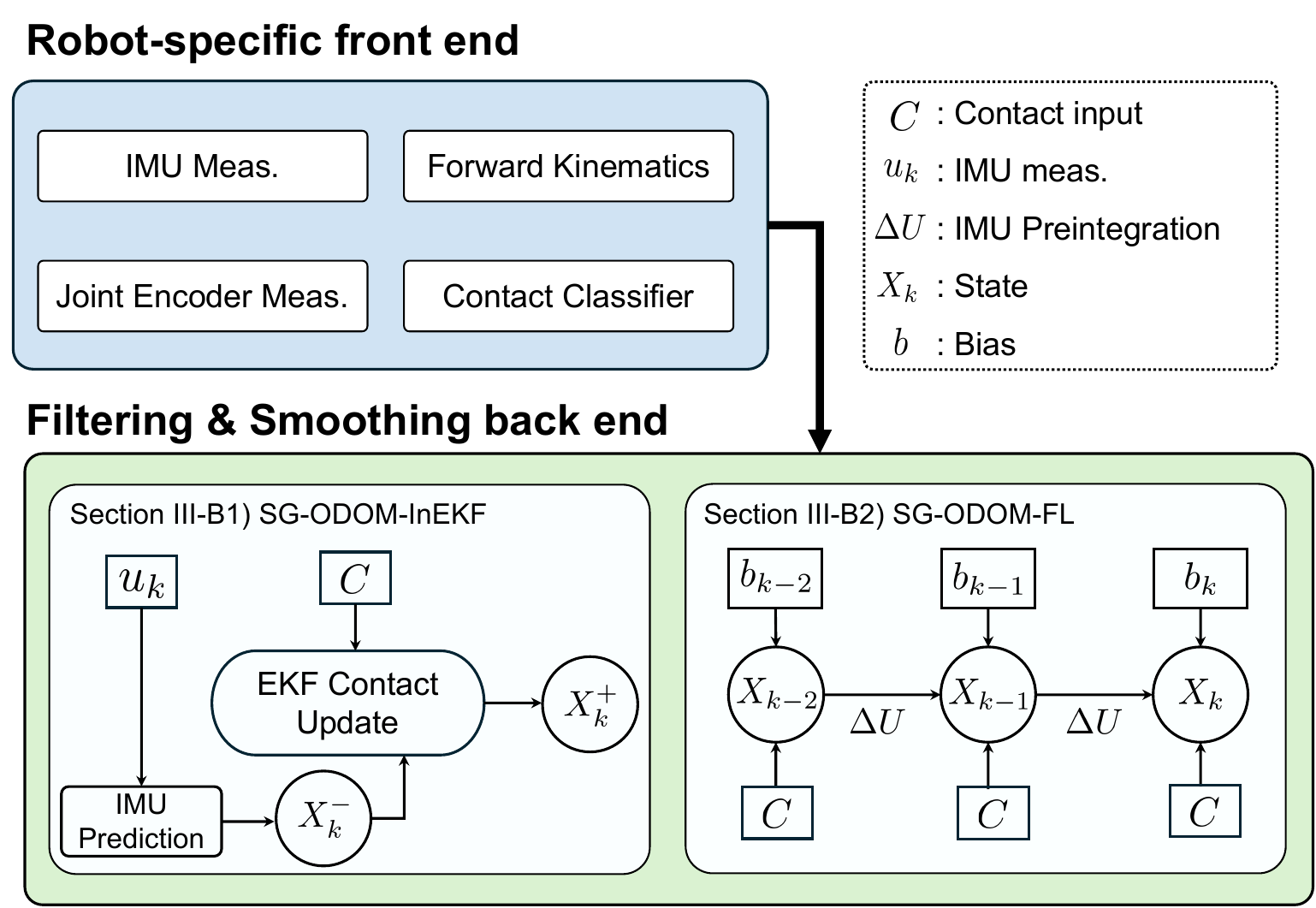}
    \caption{Overview of the proposed modular proprioceptive odometry framework. The robot-specific front end converts platform-dependent sensing and kinematics into common interfaces, which are processed by a shared estimation core supporting \ac{InEKF} and fixed-lag smoothing.}
    \label{fig:gtsam_configuration}
    \vspace{-7mm}
\end{figure}

Figure~\ref{fig:gtsam_configuration} illustrates the two configurations supported in SG-ODOM. Both share the same event-based contact-update schedule, described in detail in Sec.~\ref{sec:event-based-contact}. The two estimator configurations are included to compare recursive filtering and fixed-lag smoothing under the same SG-ODOM update strategy. 

\subsubsection{SG-ODOM-InEKF}
\label{sec:inv-ekf}

The contact-aided invariant filter~\cite{hartley2019contact} augments the floating-base state with one fixed foothold slot per foot. With $n_c$ such slots, the augmented state is represented as
\begin{equation}
    \mathcal{X}_k =
    \left(
        R^N_{B,k},
        p^N_{B,k},
        v^N_{B,k},
        f^N_1,
        \ldots,
        f^N_{n_c}
    \right)
    \in \mathrm{SE}_{n_c+2}(3).
    \label{eq:inekf-state}
\end{equation}
We use a left-invariant error convention,
\begin{equation}
    \mathcal{X}_k
    =
    \hat{\mathcal{X}}_k\mathrm{Exp}(\xi_k),
    \qquad
    \xi_k \in \mathbb{R}^{9+3n_c},
    \label{eq:left-invariant-error}
\end{equation}
where $\xi_k$ denotes the local error coordinates and $\mathrm{Exp}$ maps these coordinates to the Lie group. Footholds are stored in the navigation frame, while their local perturbations are expressed in the body frame.

The mean and covariance are propagated at the IMU rate using zero-order-hold inertial integration and the left-linear prediction formulation~\cite{Chirikjian11book_Stochastic,barrau2019linear}. Contact corrections follow the event-based update schedule described in Sec.~\ref{sec:event-based-contact}. Ended contact episodes are marginalized in place at contact updates, and the corresponding slots are reused for subsequent episodes.

\subsubsection{SG-ODOM-FL}
\label{sec:fl-single}

The fixed-lag smoother jointly optimizes event-based base states and explicit contact-episode landmarks $f^N_{i,e}$. Each update event defined by the schedule in Sec.~\ref{sec:event-based-contact} creates a new base state. Consecutive base states are connected by preintegrated IMU factors~\cite{forster2017manifold} that summarize the intervening IMU measurements, while contact factors follow~Eq.~\eqref{eq:contact-model}.
A separate IMU bias variable $b_j\in\mathbb{R}^{6}$, comprising gyroscope and accelerometer biases, is associated at each event state. Consecutive bias variables are modeled with a random-walk model, allowing the estimated inertial bias to evolve over time.  Each accepted contact episode is represented by a distinct landmark shared by the associated event states. Older states and contact episodes are marginalized as they leave the lag window. Between events, a current estimate is obtained by IMU propagation from the latest optimized base state.

\section{Contact Processing}
\label{sec:contact-validation}

\subsection{Event-Based Contact Updates}
\label{sec:event-based-contact}

Contact classification and estimator corrections are intentionally decoupled. Contact information is processed at the incoming measurement rate, while kinematic corrections are applied only at selected update events.
\begin{figure}
    \centering
    \includegraphics[width=1.0\linewidth, trim=0.1cm 3.2cm 0.1cm 3.2cm, clip]{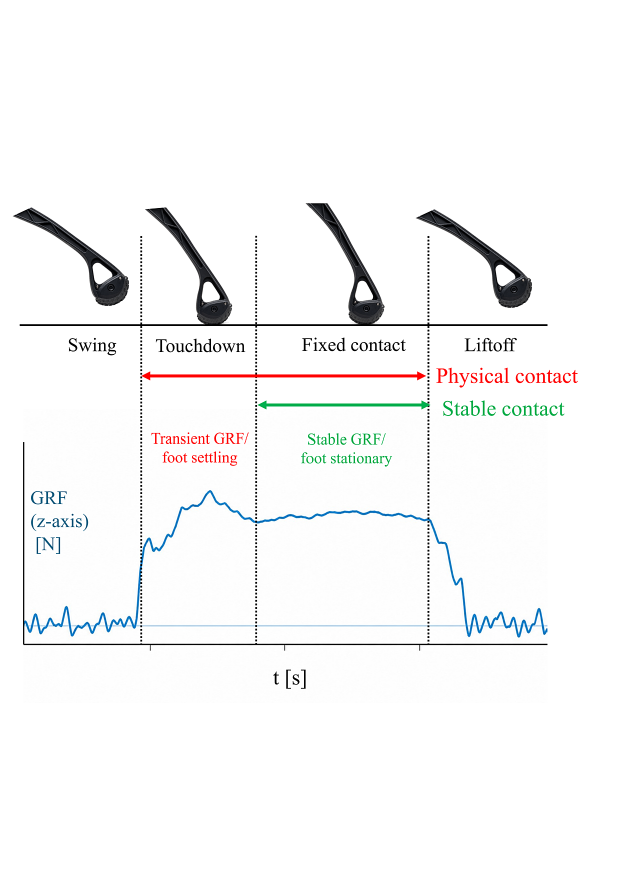}
    \caption{The highlighted foot is shown over the course of pre-touchdown swing, impact and transient loading, stable contact, and liftoff. The vertical \ac{GRF} illustrates how contact activation is delayed until the loading transient has stabilized, rather than occurring at the earliest instant of physical touchdown.}
    \label{fig:contact-classification}
    \vspace{-7mm}
\end{figure}
A newly validated touchdown immediately triggers a contact-triggered update. In the absence of a new touchdown, the estimator propagates using IMU samples until the elapsed time since the previous correction reaches the maximum allowed dead-reckoning interval, $\Delta t_{\max}$; the next valid contact measurement triggers a timeout-triggered update, even if the contact condition has not changed. Measurements received before this interval are evaluated but do not trigger an estimator correction.

Importantly, each touchdown resets the dead-reckoning interval. Therefore, as shown in Fig.~\ref{fig:event-based-contact}, timeout-triggered updates are scheduled relative to the most recent contact correction rather than to a fixed global sampling grid. For example, if a touchdown occurs before $\Delta t_{\max}$ expires, the pending timeout is discarded and a new interval starts from the touchdown event.

This strategy differs from fixed-rate downsampling: all contact measurements are processed, while only the frequency of kinematic corrections is reduced. The resulting correction times are therefore not tied to a fixed sampling grid, but instead follow the sequence of contact events. Applying kinematic corrections at a lower rate avoids repeatedly using closely spaced measurements from the same contact episode and reduces the number of estimator updates, while $\Delta t_{\max}$ limits the duration of IMU-only propagation. The effect of this trade-off on estimation accuracy is evaluated experimentally in Sec.~\ref{sec:analysis-rate}.


\subsection{Stability-Gated Contact Classification}
\label{sec:contact-classification}

As illustrated in Fig. \ref{fig:contact-classification}, the first indication of physical touchdown does not necessarily imply that the foot provides a stable reference for a kinematic correction. The objective is therefore not to identify the earliest instant of physical touchdown, but to determine when the contact becomes suitable for state estimation. We therefore distinguish physical touchdown from stable touchdown. A stable touchdown is declared only when the contact conditions remain stable for a certain amount of time. For each foot $i$, the classifier produces $\gamma_i(t) \in \{0,1\},$ where $\gamma_i(t)=1$ indicates that the contact is accepted for a kinematic update. 

\subsubsection{Force Estimation and Stability Measure}
\label{sec:contact-classifier}

For each foot, the ground reaction force is estimated from the measured joint efforts following the torque-based formulation of \cite{camurri2017probabilistic}. The model-predicted joint torque is obtained through the recursive Newton--Euler algorithm implemented in Pinocchio \cite{carpentier2019pinocchio}. The resulting torque residual is mapped to the foot through the translational Jacobian as
\begin{equation}
    \mathbf f_i
    =
    -\left(J_i^\top\right)^{-1}
    \left(
        \tau_{\mathrm{m},i}-\tau_{{\mathrm{model},i}}
    \right),
    \label{eq:grf-estimation}
\end{equation}
where $J_i$ is the translational Jacobian of foot $i$, $\tau_{m_i}$ is the measured joint torque, and $\tau_{\mathrm{model},i}$ is the corresponding model-predicted torque. Only the signed vertical component, denoted by $F_i(t)$, is retained for contact classification.

Force magnitude alone does not distinguish stable support from the transient following touchdown. We therefore also consider its variation over a fixed interval $\Delta t_F$,
\begin{equation}
\Delta F_i(t)
=
\left|
F_i(t)-F_i(t-\Delta t_F)
\right|.
\label{eq:force-delta}
\end{equation}

\subsubsection{Stability-Gated Contact State Update}

The contact state changes only when the corresponding transition condition holds persistently. Contact stability is evaluated through a stability gate that requires both sufficient loading and limited force variation, 
\begin{equation}
    s_i(t)
    =
    \big(F_i(t) \geq F_{\mathrm{on}}\big)
    \wedge
    \big(\Delta F_i(t) \leq \Delta F_{\mathrm{max}}\big),
    \label{eq:stability-predicate}
\end{equation}
and the contact flag is updated according to
\begin{equation}
    \begin{aligned}
        \gamma_i:\; 0 \rightarrow 1
        &\quad \text{if } s_i(\tau) = 1
        &&\forall \tau \in [t - T_{\mathrm{on}},\, t], \\
        \gamma_i:\; 1 \rightarrow 0
        &\quad \text{if } s_i(\tau) = 0
        &&\forall \tau \in [t - T_{\mathrm{off}},\, t].
    \end{aligned}
    \label{eq:contact-transitions}
\end{equation}

The resulting stability-gated activation does not impose a fixed delay after physical touchdown. A contact is accepted only when the loading and force-stability conditions remain satisfied continuously for $T_{\mathrm{on}}$ so the effective delay varies across steps according to how rapidly $\Delta F_i(t)$ settles after impact. Until activation, measurements from that foot are excluded from kinematic updates, while other validated contacts remain eligible.

\section{Experimental Evaluation}
\label{sec:experiments}

\begin{figure}[t]
    \centering
    \includegraphics[width=1.0\linewidth]{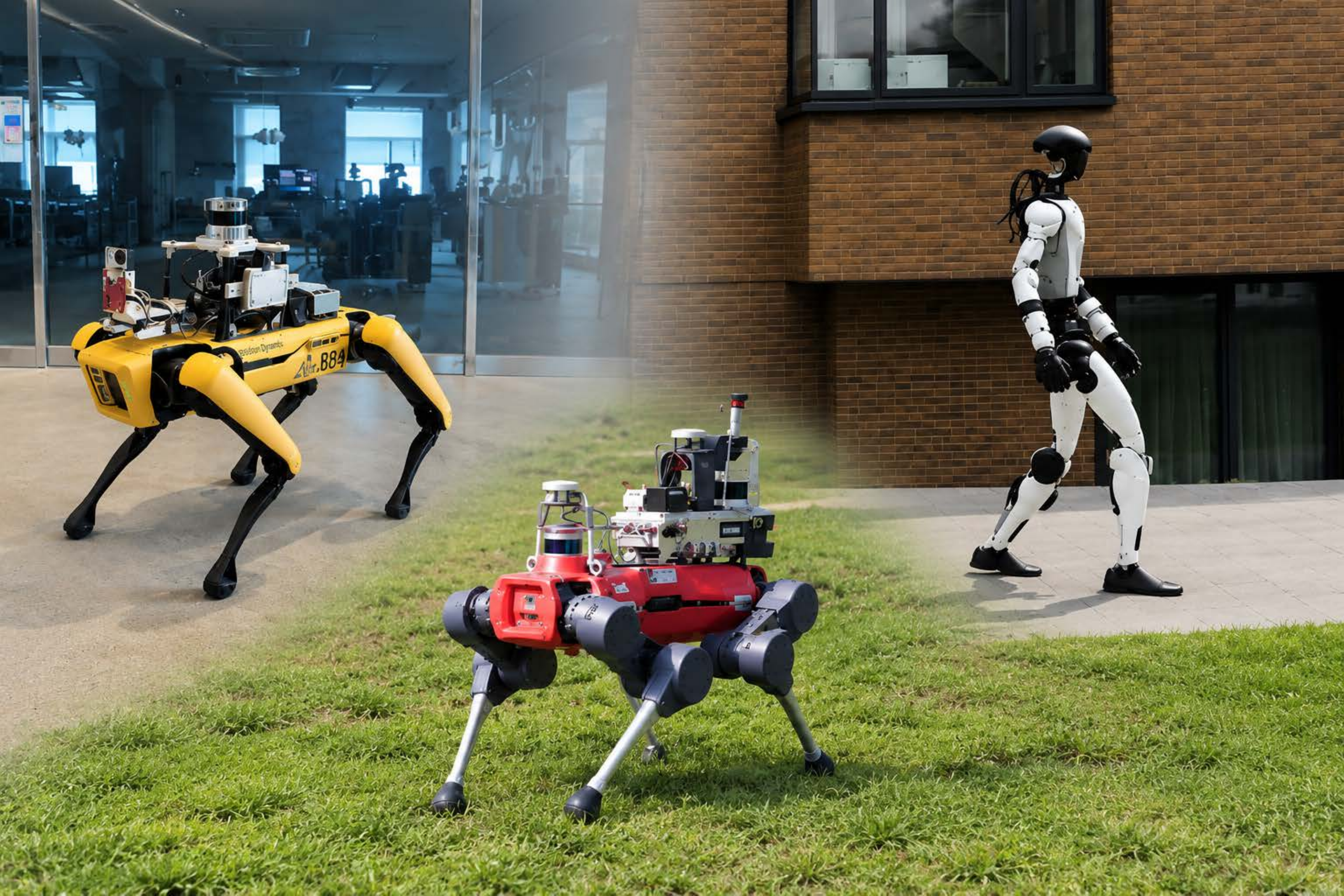}
    \caption{The proposed proprioceptive odometry framework is demonstrated across three legged platforms: Boston Dynamics Spot (left), ANYbotics ANYmal D (center), Unitree G1 humanoid (right).}
    \label{fig:cover_pic}
    \vspace{-5mm}
\end{figure}

In our experiments, we evaluate SG-ODOM across the three robotic platforms shown in Fig.~\ref{fig:cover_pic}. The two SG-ODOM configurations described in Sec.~\ref{sec:estimator-configurations} use identical IMU and joint measurements, forward kinematics, contact information, and event-driven update schedule. For each sequence, the estimated body trajectory is compared with the reference trajectory using the \ac{RMSE} of translational and rotational \ac{RPE}, in meters and degrees, respectively. \ac{RPE} is computed over \(1\,\mathrm{m}\) intervals. 

Both SG-ODOM estimator configurations are implemented in GTSAM~\cite{gtsam}. The SG-ODOM-InEKF uses its filtering infrastructure, while the SG-ODOM-FL uses IMU preintegration, factor-graph optimization, and fixed-lag marginalization.

In Sec.~\ref{sec:analysis-rate} we assess how the contact update interval affects  estimator accuracy. Sec.~\ref{sec:contact-validation-experiment} assesses the effect of the stability-gated contact classification.

\paragraph{SG-ODOM-FL Bias Random-Walk Tuning}
For SG-ODOM-FL, the gyroscope- and accelerometer-bias random-walk parameters were selected independently for each robotic platform and then kept fixed across its evaluation sequences. Dedicated IMU calibration data is not available for the three platforms; as a result, the parameters were selected empirically rather than derived from a common Allan-deviation analysis. These parameters control how rapidly estimated inertial biases are allowed to evolve within the fixed-lag window.

\paragraph{Datasets}
We evaluate ANYmal on seven outdoor missions from the GrandTour
dataset~\cite{frey_tuna2026grandtour}: ALB-3, GRI-1, HEAP-1, LEICA-1,
PIL-2, SBB-1, and SNOW-3, covering forest, urban, industrial,
mountainous, and snow-covered terrain.
Spot is evaluated on four indoor and outdoor sequences from the
GaRLILEO dataset~\cite{noh2026garlileo}: DOWNSTAIR, OVERPASS,
MOCAP-E, and MOCAP-H.
Unitree G1 is evaluated on two self-collected outdoor sequences:
QUAD, a \(112.5\,\mathrm{m}\) paved courtyard loop, and SLOPE,
a \(64.4\,\mathrm{m}\) ascent and descent of a \(4^\circ\) slope.

\subsection{Analysis of the Maximum Dead-Reckoning Interval}
\label{sec:analysis-rate}
A central question in the proposed formulation is how frequently kinematic corrections are required. We therefore study how estimation accuracy is affected by $\Delta t_{\max}$, which defines the maximum interval between successive kinematic corrections in the absence of a new contact-triggered update. For each SG-ODOM configuration, we vary $\Delta t_{\max}$ while keeping the contact classifier and all remaining parameters unchanged, and report translational RPE in Table~\ref{tab:rate-comparison-inekf} and Table~\ref{tab:rate-comparison-fl}. We focus this ablation on translational RPE.

$\Delta t_{\max}$ was varied between $0.01s$ to $0.30s$. For the sequences considered in this evaluation, the observed gait periods are approximately $0.65,\mathrm{s}$ for ANYmal, $0.70,\mathrm{s}$ for Spot, and $0.85,\mathrm{s}$ for Unitree G1. The results exhibit a clear intermediate region with the lowest error, with nearly identical performance at $0.05,\mathrm{s}$ and $0.10,\mathrm{s}$. For the quadruped platforms, ANYmal and Spot $\Delta t_{\max} = 0.10,\mathrm{s}$ corresponds to approximately three to four kinematic corrections during a typical foot-contact phase. When $\Delta t_{\max}$ becomes large relative to the gait cycle, the updates become too sparse to constrain inertial drift between contact events. For SG-ODOM-FL, very short intervals also increase the error, particularly on ANYmal and Spot. In contrast, SG-ODOM-InEKF remains comparatively stable as the update interval is reduced, indicating that increasing the frequency of kinematic corrections does not provide a corresponding reduction in estimation error. 

At short update intervals, consecutive joint measurements may provide limited additional information because they are strongly correlated, while residual foot motion can still violate the stationary-contact assumption. This behavior is less pronounced on the humanoid robot, with the profile being much flatter across the range of $\Delta t_{\max}$. This may be due to the G1's slower, smoother walking gait on paved terrain. Based on this analysis, we adopt $\Delta t_{\max} = 0.10s$ for all our remaining experiments.

\begin{table}[!t]
    \centering
    \caption{Effect of the maximum interval between kinematic corrections
    on translational RPE for SG-ODOM-InEKF.
    RPE$_t$ is reported in meters, and $T_{\mathrm{gait}}$ denotes the average gait period for each platform.}
    \label{tab:rate-comparison-inekf}
    \small
    \setlength{\tabcolsep}{7pt}
    \renewcommand{\arraystretch}{1.0}
    \begin{tabular}{l|c|c|c}
        \hline
        \multirow{2}{*}{$\Delta t_{\max}$} & ANYmal & Spot & Unitree G1 \\
         & {\scriptsize $(T_{\mathrm{gait}}=0.65\,\mathrm{s})$}
        & {\scriptsize $(T_{\mathrm{gait}}=0.70\,\mathrm{s})$}
        & {\scriptsize $(T_{\mathrm{gait}}=0.85\,\mathrm{s})$} \\
        \hline
        $0.30\,\mathrm{s}$ & 0.484 & 0.512 & 0.131 \\
        $0.20\,\mathrm{s}$ & 0.265 & 0.227 & 0.119 \\
        $0.10\,\mathrm{s}$ & \underline{0.252} & \textbf{0.210} & \underline{0.115} \\
        $0.05\,\mathrm{s}$ & \textbf{0.250} & \underline{0.211} & \textbf{0.114} \\
        $0.01\,\mathrm{s}$ & {0.255} & 0.213 & 0.122 \\
        \hline
    \end{tabular}
\end{table}

\begin{table}[!t]
    \centering
    \caption{Effect of the maximum interval between kinematic corrections
    on translational RPE for SG-ODOM-FL.
    RPE$_t$ is reported in meters, and $T_{\mathrm{gait}}$ denotes the average gait period for each platform.}
    \label{tab:rate-comparison-fl}
    \small
    \setlength{\tabcolsep}{7pt}
    \renewcommand{\arraystretch}{1.0}
    \begin{tabular}{l|c|c|c}
        \hline
        \multirow{2}{*}{$\Delta t_{\max}$} & ANYmal & Spot & Unitree G1 \\
        & {\scriptsize $(T_{\mathrm{gait}}=0.65\,\mathrm{s})$}
        & {\scriptsize $(T_{\mathrm{gait}}=0.70\,\mathrm{s})$}
        & {\scriptsize $(T_{\mathrm{gait}}=0.85\,\mathrm{s})$} \\
        \hline
        $0.30\,\mathrm{s}$ & 0.418 & 0.504 & 0.121 \\
        $0.20\,\mathrm{s}$ & 0.202 & 0.215 & 0.114 \\
        $0.10\,\mathrm{s}$ & \textbf{0.157} & \textbf{0.185} & \textbf{0.109} \\
        $0.05\,\mathrm{s}$ & \textbf{0.157} & \underline{0.197} & \underline{0.110} \\
        $0.01\,\mathrm{s}$ & 0.406 & 0.398 & 0.118 \\
        \hline
    \end{tabular}
    \vspace{-3mm}
\end{table}

\subsection{Analysis of the Contact Classification Algorithm}
\label{sec:contact-validation-experiment}
\paragraph{Contact Behavior Analysis}

\begin{figure}
    \centering
    \includegraphics[width=1.0\columnwidth,trim=0.3cm 0.2cm 0.1cm 0.1cm,
            clip ]{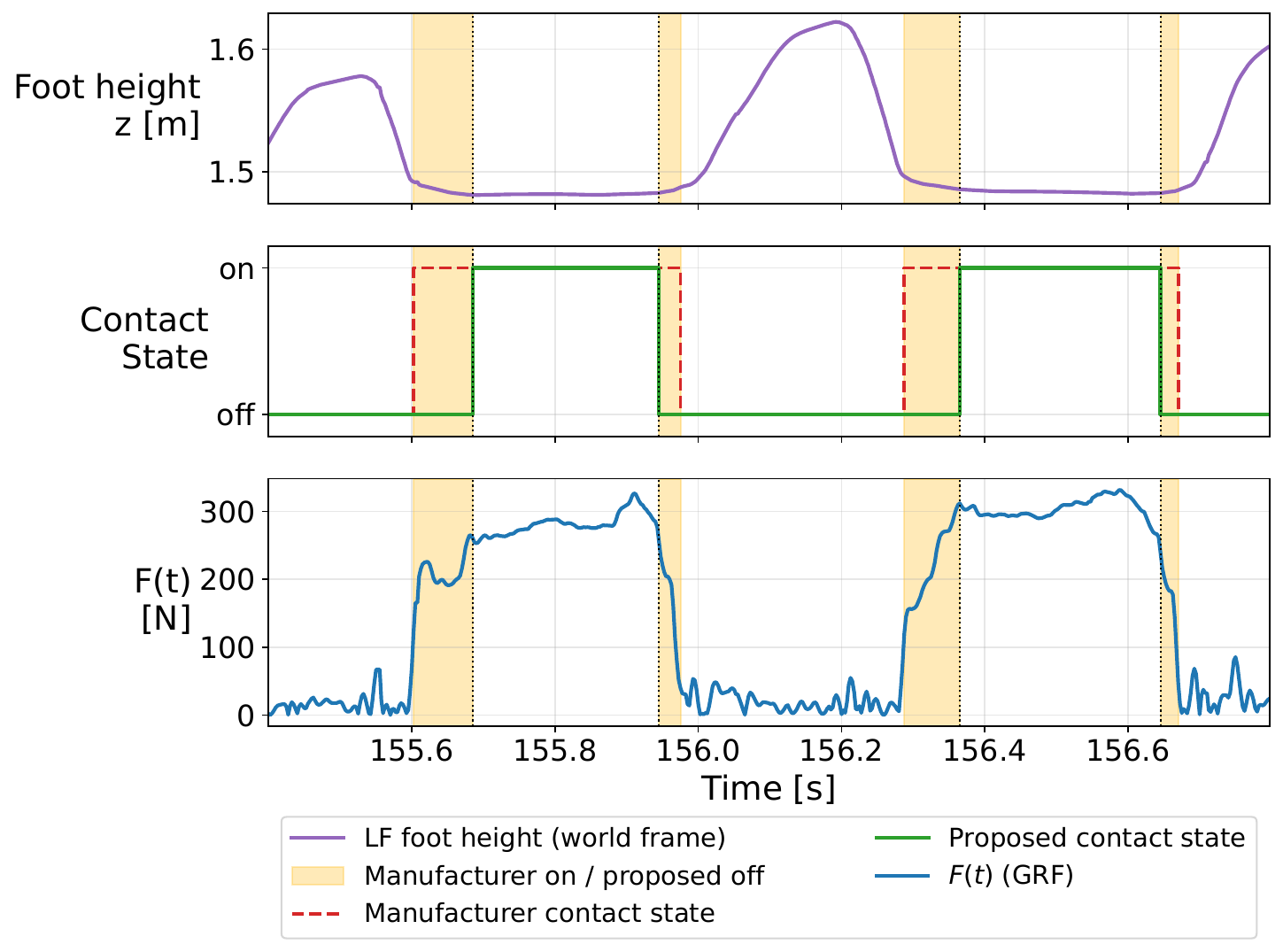}
    \caption{Contact classification for the left-front foot of ANYmal while trotting. Top to bottom: world-frame foot height, contact classification by ANYbotics (red) classifier and the implemented one $\gamma_i(t)$, and estimated vertical \ac{GRF}. Highlighted intervals mark samples accepted by the default classifier but rejected by the $\gamma_i(t)$ during contact transients.}
    \label{fig:contact-validation}
    \vspace{-3mm}
\end{figure}

We analyze the behavior of the proposed stability-gated contact classifier using the ANYbotics ANYmal D with the HEAP-1 sequence from the GrandTour dataset. Fig.~\ref{fig:contact-validation} shows a representative contact event for its left-front foot. The upper panel reports the reconstructed world-frame foot height, the middle panel compares the default ANYbotics contact state with the implemented contact state $\gamma_i(t)$, and the lower panel shows the estimated vertical \ac{GRF} $F_{i}(t)$. The ground-truth foot position was obtained by combining the ground-truth base pose with the robot's joint sensing and is used only for analysis, not for contact classification.

Fig.~\ref{fig:contact-validation} shows that contact samples selected by the default ANYbotics classifier near loading and unloading coincide with continued odom-frame foot motion and pronounced changes in the estimated vertical \ac{GRF}. The proposed stability-gated classifier rejects these transient portions and accepts contact only after the stability conditions defined in Sec.~\ref{sec:contact-classification} remain satisfied for the required validation interval. The manufacturer-provided classifier may be tuned for different objectives beyond odometry accuracy, and we are unaware of the specific considerations behind its tuning.

\paragraph{Effect of Contact Classification Strategy}
\label{sec:contact-effect}
\begin{table}[!t]
    \centering
    \caption{Effect of the contact strategy on translational RPE for SG-ODOM-InEKF. RPE$_t$ is reported in meters. Bold values indicate the best results.}
    \label{tab:contact-ablation-inekf}
    \small
    \setlength{\tabcolsep}{7pt}
    \renewcommand{\arraystretch}{1.0}
    \begin{tabular}{l|c|c|c}
        \hline
        Contact strategy & ANYmal & Spot & Unitree G1 \\
        \hline
        Manufacturer-provided & 0.291 & 0.224 & -- \\
        Force threshold        & 0.307 & 0.228 & 0.122 \\
        Stability-gated        & \textbf{0.252} & \textbf{0.210} & \textbf{0.115} \\
        \hline
    \end{tabular}
    \vspace{-2mm}
\end{table}

\begin{table}[!t]
    \centering
    \caption{Effect of the contact strategy on translational RPE for SG-ODOM-FL. RPE$_t$ is reported in meters. Bold values indicate the best  results.}
    \label{tab:contact-ablation-fl}
    \small
    \setlength{\tabcolsep}{7pt}
    \renewcommand{\arraystretch}{1.0}
    \begin{tabular}{l|c|c|c}
        \hline
        Contact strategy & ANYmal & Spot & Unitree G1 \\
        \hline
        Manufacturer-provided & {0.210} & 0.203 & -- \\
        Force threshold        & 0.259 & {0.190} & {0.113} \\
        Stability-gated        & \textbf{0.157} & \textbf{0.185} & \textbf{0.109} \\
        \hline
    \end{tabular}
    \vspace{-6mm}
\end{table}

We next isolate the effect of contact handling for both SG-ODOM-InEKF and SG-ODOM-FL while varying only the contact strategy. Translational RPE is used as the common metric in all experiments with the three platforms. The proposed stability-gated contact classifier is compared against a simple force-threshold strategy, with the manufacturer-provided contact state included as an additional baseline when available.

As shown in Tables~\ref{tab:contact-ablation-inekf} and \ref{tab:contact-ablation-fl}, the stability-gated classifier achieves the lowest error for both estimator configurations on all three platforms. The largest improvement is observed on ANYmal, which traverses a wider range of natural terrain. A smaller but consistent improvement is also observed on Spot, evaluated on challenging indoor and outdoor sequences that include interaction with soft terrain. The difference is more limited on Unitree G1, which was tested only on paved concrete where the walking gait was slower and gentler.

\subsection{Comparison using Quadrupeds and Humanoids}
\label{sec:estimator-comparison}

\paragraph{ANYmal}
\begin{table*}[t]
    \centering
    \caption{Translational \ac{RPE} on ANYmal using missions from the GrandTour dataset~\cite{frey_tuna2026grandtour}. RPE$_t$ is evaluated over \(1\,\mathrm{m}\) path-length intervals and reported in meters. The final column reports the average across the seven missions. Bold and underlined values indicate the best and second-best results, respectively.}
    \label{tab:anymal-results}
    \small
    \setlength{\tabcolsep}{5pt}
    \renewcommand{\arraystretch}{1.0}
    \begin{tabular}{l|c|c|c|c|c|c|c!{\vrule width 1.0pt}c}
        \hline
        & ALB-3 & GRI-1 & HEAP-1 & LEICA-1 & PIL-2 & SBB-1 & SNOW-3
        & \textbf{Average} \\
        \hline
        MUSE~\cite{nistico2025muse}
        & 0.418 & 0.395 & 0.175 & 0.223 & 0.183 & 0.160 & 0.511 & 0.295 \\
        IEKF~\cite{hartley2019contact}
        & 0.398 & 0.401 & 0.181 & 0.194 & 0.159 & 0.360 & 0.512 & 0.315 \\
        IS~\cite{yoon2024invariant}
        & 0.397 & 0.403 & 0.182 & 0.193 & 0.158 & 0.363 & 0.515 & 0.316 \\
        ANYbotics Estimator
        & 0.366 & \underline{0.164} & \textbf{0.144} & \underline{0.155} & 0.179
        & 0.293 & \underline{0.156} & \underline{0.208} \\
        \hline
        SG-ODOM-InEKF
        & \underline{0.308} & \textbf{0.158} & 0.239 & 0.587
        & \underline{0.155} & \underline{0.149} & 0.167 & 0.252 \\
        SG-ODOM-FL
        & \textbf{0.263} & 0.169 & \underline{0.153} & \textbf{0.131}
        & \textbf{0.127} & \textbf{0.121} & \textbf{0.133}
        & \textbf{0.157} \\
        \hline
    \end{tabular}
    \vspace{-2mm}
\end{table*}

Table~\ref{tab:anymal-results} compares two SG-ODOM configurations with MUSE~\cite{nistico2025muse}, IEKF~\cite{hartley2019contact}, IS~\cite{yoon2024invariant}, and the onboard ANYbotics estimator~\cite{bloesch2013state}. The ground truth trajectory is provided by the GrandTour dataset~\cite{frey_tuna2026grandtour}, using a Leica Geosystems MS60 Total Station. For the ANYmal experiments only RPE$_t$ is reported because the provided reference trajectories are of position but not orientation.

SG-ODOM-FL achieves the lowest average RPE$_t$ of \(0.157\,\mathrm{m}\) and performs best on 5 of the 7 sequences, while the ANYbotics estimator remains best on HEAP-1. SG-ODOM-FL also shows more consistent performance across missions than SG-ODOM-InEKF and the compared baselines.

\paragraph{Spot}
\begin{table*}[t]
    \centering
    \caption{Relative pose errors on Spot using sequences from the GaRLILEO dataset~\cite{noh2026garlileo}. RPE$_t$ and RPE$_r$ are evaluated over \(1\,\mathrm{m}\) path-length intervals and reported in meters and degrees, respectively. The final column reports the average across all sequences. Bold and underlined values indicate the best and second-best results, respectively.}
    \label{tab:spot-results}
    \small
    \setlength{\tabcolsep}{5pt}
    \renewcommand{\arraystretch}{1.0}
    \begin{tabular}{l|cc|cc|cc|cc!{\vrule width 1.0pt}cc}
        \hline
        & \multicolumn{2}{c|}{Downstair}
        & \multicolumn{2}{c|}{Overpass}
        & \multicolumn{2}{c|}{MoCap-E}
        & \multicolumn{2}{c!{\vrule width 1.0pt}}{MoCap-H}
        & \multicolumn{2}{c}{\textbf{Average}} \\

        \cline{2-11}
        & RPE$_t$ & RPE$_r$ & RPE$_t$ & RPE$_r$
        & RPE$_t$ & RPE$_r$ & RPE$_t$ & RPE$_r$
        & RPE$_t$ & RPE$_r$ \\
        \hline
        Pronto~\cite{camurri2020pronto}
        & 0.196 & 0.925
        & 0.180 & \textbf{0.804}
        & 0.264 & \underline{0.652}
        & \underline{0.263} & \underline{0.887}
        & 0.226 & \underline{0.817} \\

        MUSE~\cite{nistico2025muse}
        & 0.182 & 1.095
        & 0.191 & 0.972
        & \underline{0.252} & 1.831
        & 0.267 & 1.733
        & 0.223 & 1.408 \\
        \hline

        SG-ODOM-InEKF
        & \underline{0.175} & \textbf{0.791}
        & \underline{0.125} & \underline{0.930}
        & \underline{0.252} & \textbf{0.648}
        & 0.288 & 0.888
        & \underline{0.210} & \textbf{0.814} \\

        SG-ODOM-FL
        & \textbf{0.117} & \underline{0.813}
        & \textbf{0.118} & \underline{0.930}
        & \textbf{0.250} & 0.674
        & \textbf{0.257} & \textbf{0.869}
        & \textbf{0.185} & 0.821 \\
        \hline
    \end{tabular}
    \vspace{-4mm}
\end{table*}

Table~\ref{tab:spot-results} compares the two SG-ODOM configurations with Pronto~\cite{camurri2020pronto} and MUSE~\cite{nistico2025muse}. For DOWNSTAIR and OVERPASS, the reference trajectories were generated using prior-map-assisted LiDAR localization against a high-accuracy terrestrial laser scan acquired with a Leica RTC360. MOCAP-E and MOCAP-H are compared to a Vicon motion-capture system.

SG-ODOM-FL achieves the lowest RPE$_t$ on all four sequences, averaging \(0.185\,\mathrm{m}\), approximately 18\% below Pronto. SG-ODOM-InEKF has the lowest average RPE$_r$ (\(0.814^\circ\)), although rotational accuracy remains similar across the two SG-ODOM configurations and Pronto.

\paragraph{Unitree G1}
\begin{table}[t]
    \centering
    \caption{Relative pose errors on Unitree G1 using self-collected sequences. RPE is evaluated over \(1\,\mathrm{m}\) path-length intervals and reported in meters and degrees. The final column reports the average across all sequences. Bold and underlined values indicate the best and second-best results, respectively.}
    \label{tab:unitree-results}
    \footnotesize
    \setlength{\tabcolsep}{3pt}
    \renewcommand{\arraystretch}{1.0}
    \begin{tabular}{l|cc|cc!{\vrule width 1.0pt}cc}
        \hline
        & \multicolumn{2}{c|}{QUAD}
        & \multicolumn{2}{c!{\vrule width 1.0pt}}{SLOPE}
        & \multicolumn{2}{c}{\textbf{Average}} \\
        \cline{2-7}
        & RPE$_t$ & RPE$_r$
        & RPE$_t$ & RPE$_r$
        & RPE$_t$ & RPE$_r$ \\
        \hline
        Pronto~\cite{camurri2020pronto}
        & 0.220 & 6.032 & 0.203 & 5.156 & 0.212 & 5.594 \\
        Unitree Odometry
        & 0.263 & 6.107 & 0.232 & 5.773 & 0.248 & 5.940 \\
        \hline
        SG-ODOM-InEKF
        & \underline{0.116} & \underline{5.412}
        & \underline{0.114} & \underline{2.386}
        & \underline{0.115} & \underline{3.899} \\
        SG-ODOM-FL
        & \textbf{0.112} & \textbf{5.391}
        & \textbf{0.106} & \textbf{2.328}
        & \textbf{0.109} & \textbf{3.859} \\
        \hline
    \end{tabular}
    \vspace{-5mm}
\end{table}

Table~\ref{tab:unitree-results} compares the two SG-ODOM configurations with Pronto~\cite{camurri2020pronto} and the onboard Unitree odometry. Since external tracking was unavailable during data collection, Surfel-LIO~\cite{choi2025surfellio} was used to create a reference trajectory. These results therefore measure agreement with an independent LiDAR--inertial odometry estimate rather than absolute ground truth. 

The two SG-ODOM configurations perform similarly, with SG-ODOM-FL achieving the lowest average RPE$_t$ (\(0.109\,\mathrm{m}\)), approximately half that of either baseline. Rotational errors for SG-ODOM-FL are comparable to Pronto on QUAD and less than half its error on SLOPE.

\subsection{Representative Trajectories}
\label{sec:representative-trajectories}

\begin{figure*}
    \centering

    \begin{subfigure}[t]{0.32\textwidth}
        \centering
        \includegraphics[width=\linewidth]{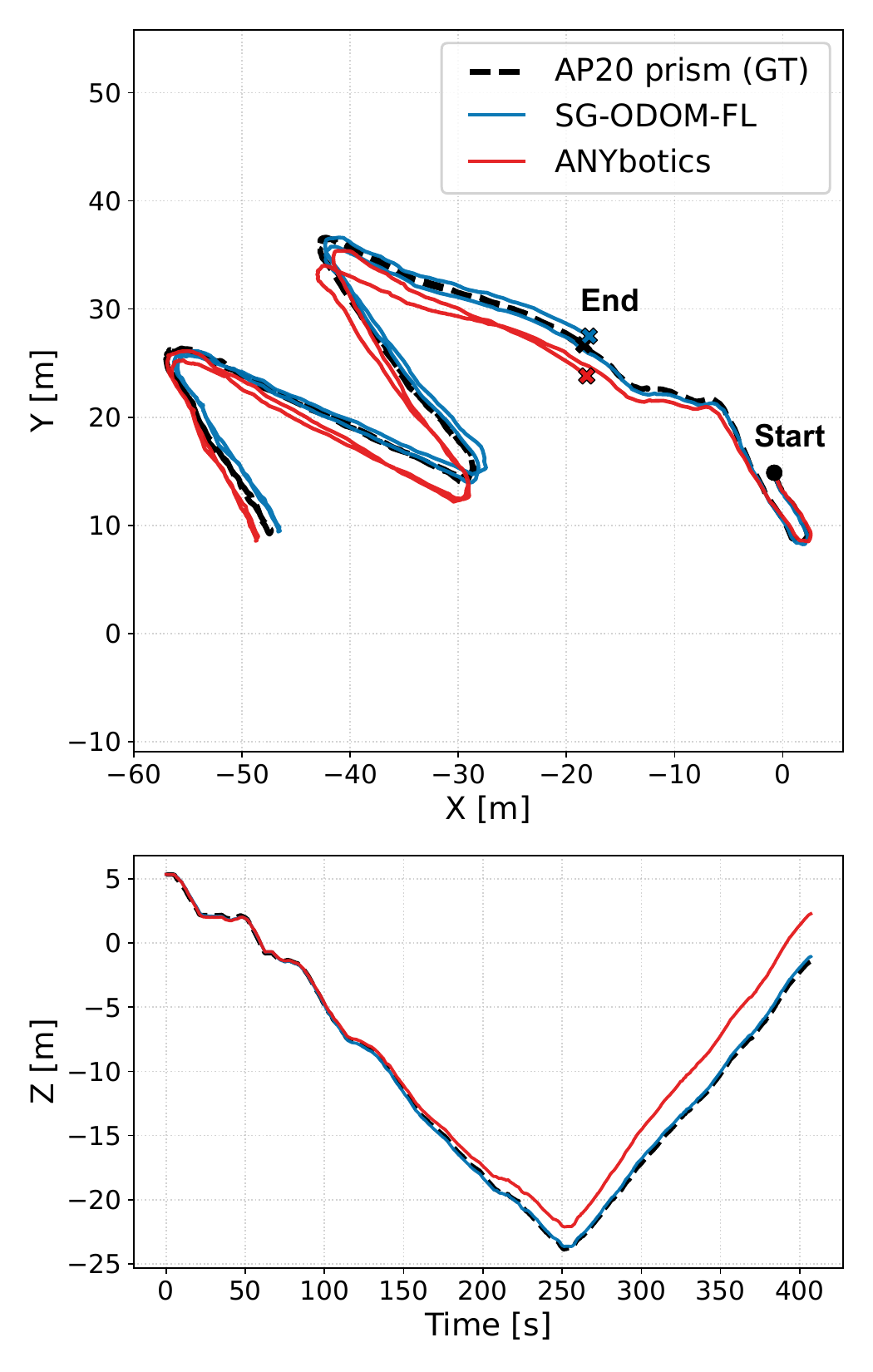}
        \caption{ANYmal on PIL-2}
        \label{fig:sub1}
    \end{subfigure}
    \hfill
    \begin{subfigure}[t]{0.32\textwidth}
        \centering
        \includegraphics[width=\linewidth]{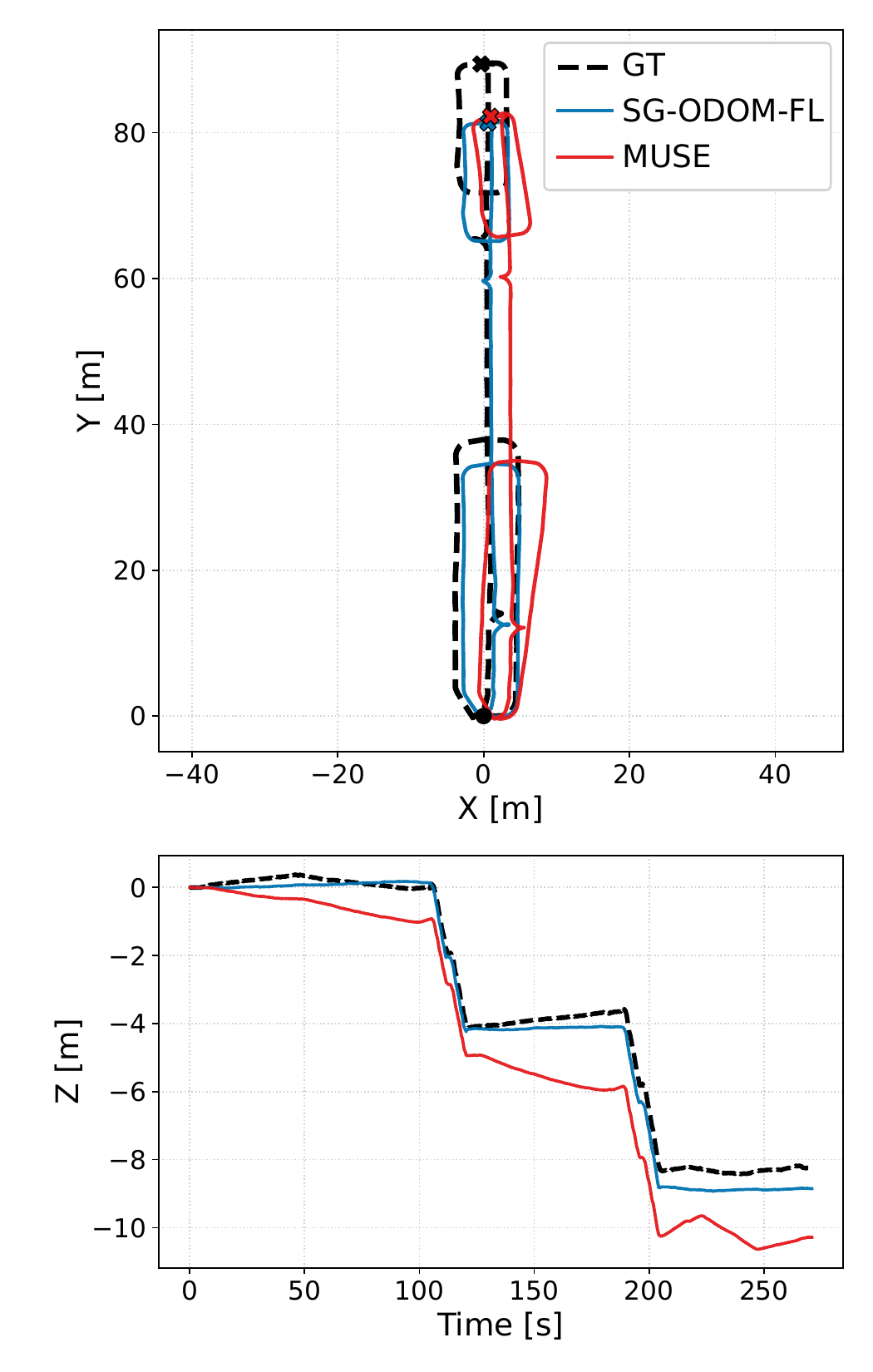}
        \caption{Spot on DOWNSTAIR}
        \label{fig:sub2}
    \end{subfigure}
    \hfill
    \begin{subfigure}[t]{0.32\textwidth}
        \centering
        \includegraphics[width=\linewidth]{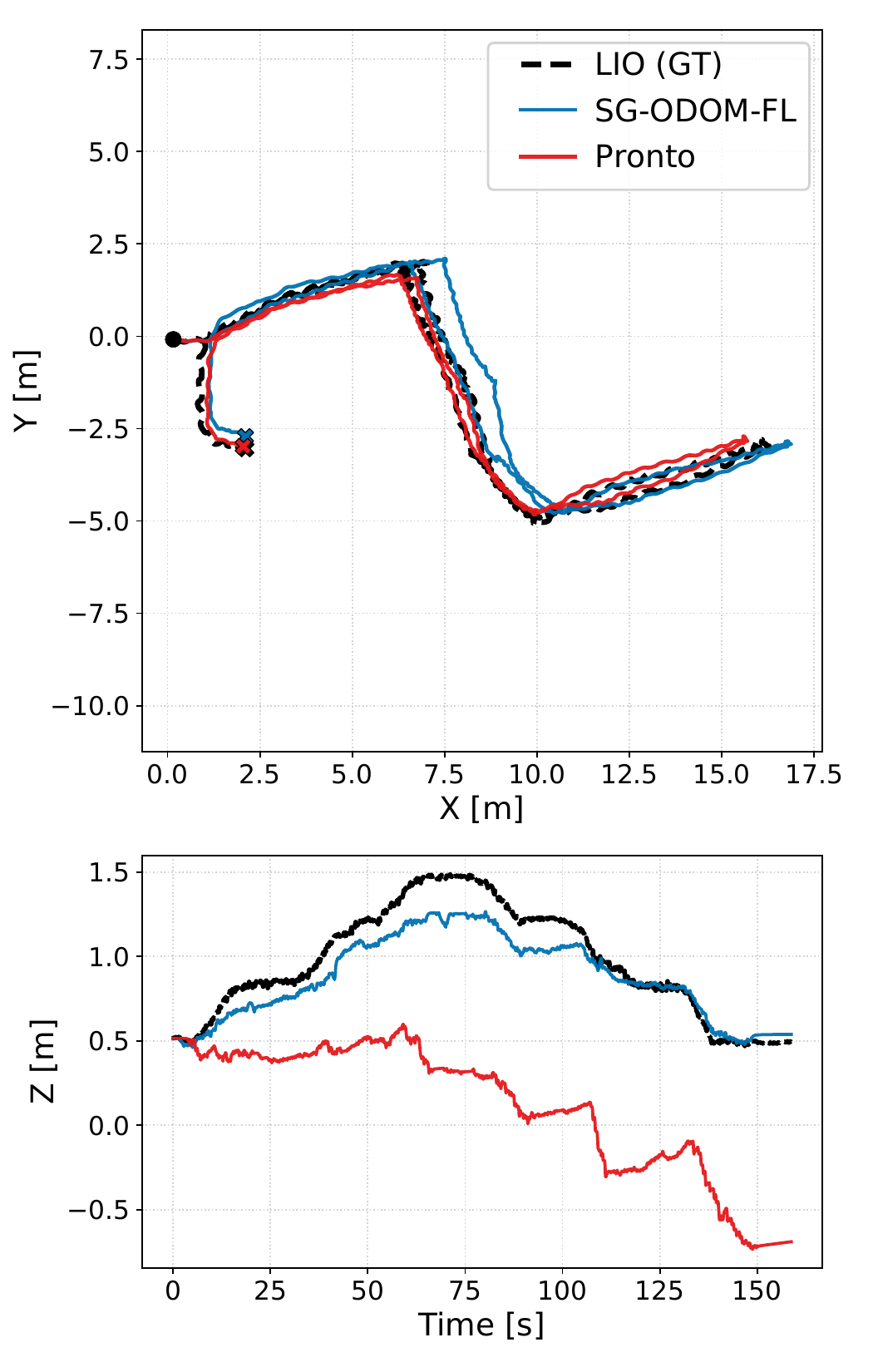}
        \caption{Unitree G1 on SLOPE}
        \label{fig:sub3}
    \end{subfigure}

    \caption{Representative trajectory estimates. Top row shows trajectories in horizontal plane and bottom row shows vertical trajectories. Circles denote start points, and crosses denote end points.}
    \label{fig:trajectories}
    \vspace{-5mm}
\end{figure*}

Figure~\ref{fig:trajectories} illustrates representative estimates for ANYmal on PIL-2, Spot on DOWNSTAIR, and Unitree G1 on SLOPE. Each example compares SG-ODOM-FL with a baseline method. On PIL-2, Fig. \ref{fig:sub1}, SG-ODOM-FL shows smaller horizontal and vertical discrepancies than the ANYbotics estimator. On DOWNSTAIR, Fig. \ref{fig:sub2}, SG-ODOM-FL follows the staircase descents more closely than MUSE, with less accumulated vertical drift. On SLOPE, Fig. \ref{fig:sub3}, both SG-ODOM-FL and Pronto reproduce the horizontal route, but Pronto accumulates downward drift while SG-ODOM-FL follows the reference elevation more closely.


\section{Conclusion}
\label{sec:conclusion}




We presented an event-driven framework for proprioceptive legged state estimation that carries out high-rate IMU propagation and contact classification, while using stability-gating to trigger kinematic updates. The strategy was implemented as an \ac{InEKF} and a fixed-lag smoother, with shared sensing, contact, and update interfaces. We evaluated on two quadrupedal robots and a humanoid, showing that both contact validation and update timing affect estimation accuracy:
long update intervals degraded both configurations; very short intervals degraded the smoother on the quadruped datasets.  Stability-gated contact updates improved translational accuracy by avoiding transient contact phases.

Our results support decoupling high-rate proprioceptive propagation and contact classification from the timing of kinematic corrections. Future work will focus on adaptive contact-validation criteria and tighter integration with complementary sensing modalities.
\section*{ACKNOWLEDGEMENTS}

{
ChatGPT was used to assist with manuscript drafting, language editing, and image editing in Fig.~4. The authors reviewed and approved all resulting content and take full responsibility for the final manuscript.
}

\small
\bibliographystyle{IEEEtranN}
\bibliography{references}

\end{document}